\documentclass[letterpaper, 10 pt, conference]{ieeeconf}  % Comment this line out if you need a4paper

\IEEEoverridecommandlockouts                              % This command is only needed if 
\usepackage{multicol}
\usepackage[bookmarks=true]{hyperref}
\usepackage{graphicx}
\usepackage{amsmath}
\usepackage{amssymb}
\graphicspath{{./images/}}
\usepackage[linesnumbered,ruled,vlined]{algorithm2e} 

\SetCommentSty{mycommfont}
\usepackage{authblk}
\usepackage{xcolor}
\usepackage{comment}
\usepackage{diagbox}
\usepackage{wrapfig}
\usepackage{makecell}
\usepackage{arydshln}

\usepackage{multirow}

\title{\LARGE \bf Adversarial Imitation Learning with Trajectorial Augmentation and Correction}

\author{Dafni Antotsiou, Carlo Ciliberto and Tae-Kyun Kim% <-this % stops a space
\thanks{The authors are with the EEE Department, Imperial College London, UK. {\tt \{d.antotsiou17, c.ciliberto, tk.kim\}@imperial.ac.uk}}% <-this % stops a space
\thanks{Project website: \url{https://sites.google.com/view/cat-dauggi}}
}

\begin{document}

\maketitle
\thispagestyle{empty}
\pagestyle{empty}

%%%%%%%%%%%%%%%%%%%%%%%%%%%%%%%%%%%%%%%%%%%%%%%%%%%%%%%%%%%%%%%%%%%%%%%%%%%%%%%%
\begin{abstract}
Deep Imitation Learning requires a large number of expert demonstrations, which are not always easy to obtain, especially for complex tasks. A way to overcome this shortage of labels is through data augmentation. However, this cannot be easily applied to control tasks due to the sequential nature of the problem. In this work, we introduce a novel augmentation method which preserves the success of the augmented trajectories. To achieve this, we introduce a semi-supervised correction network that aims to correct distorted expert actions. To adequately test the abilities of the correction network, we develop an adversarial data augmented imitation architecture to train an imitation agent using synthetic experts. Additionally, we introduce a metric to measure diversity in trajectory datasets. Experiments show that our data augmentation strategy can improve accuracy and convergence time of adversarial imitation while preserving the diversity between the generated and real trajectories.
\end{abstract}

%%%%%%%%%%%%%%%%%%%%%%%%%%%%%%%%%%%%%%%%%%%%%%%%%%%%%%%%%%%%%%%%%%%%%%%%%%%%%%%%
\section{INTRODUCTION}
Imitation learning (IL) leverages sample demonstrations from an expert to train an autonomous agent on a variety of complex tasks~\cite{hussein2017imitation}. The main advantages of learning from demonstrations are a more ``natural'' behaviour for the trained system and no need to design task-specific hand-crafted features. This is especially favourable when compared with Reinforcement Learning (RL)~\cite{sutton2018reinforcement} strategies, which, in contrast, require a well-designed reward function per task. In an effort to combine RL elements with supervised learning, generative adversarial imitation learning (GAIL)~\cite{ho2016generative} developed an adversarial imitation architecture, where a generator competes with a discriminator to match the distribution of the experts. While this strategy has obtained promising results, it still requires a large dataset of diverse expert trajectories. This process is often challenging, for instance due to the recording device, as well as the fact that retargeting to the robot domain often introduces substantial noise~\cite{antotsiou2018task}. Additionally, expert recordings often require bespoke equipment~\cite{Rajeswaran-RSS-18}, making the recording process very expensive and time-consuming. This ends up being a major bottleneck for most real-world applications, especially in deep learning settings, where a large number of demonstrations is typically needed~\cite{qi2019small}.
\par
In this work we propose a novel strategy to learn to generate synthetic experts from few examples. Our goal is to reduce the impact of input noise and generalise to different conditions when the original dataset size is limited. Contrary to other data augmentation approaches for control problems~\cite{bojarski2016end,george2018imitation,buhet2019conditional}, we devise a system that performs random trajectory augmentation. In order to ensure the success of this randomisation, we introduce a correction policy which aims to correct these random augmentations thanks to its adversarial architecture. This policy can generate potentially infinite synthetic experts, which can be used to train an imitation agent. However, to ensure the validity of the synthetic experts, our method utilises a binary success filter at the end of each trajectory. Whilst this information is not part of pure imitation, it is easy to obtain, but also not enough to successfully train an RL agent, as we empirically demonstrate in Section~\ref{sec:exp}.
\par
\begin{figure}[t]
\centering
\includegraphics[width=0.46\textwidth]{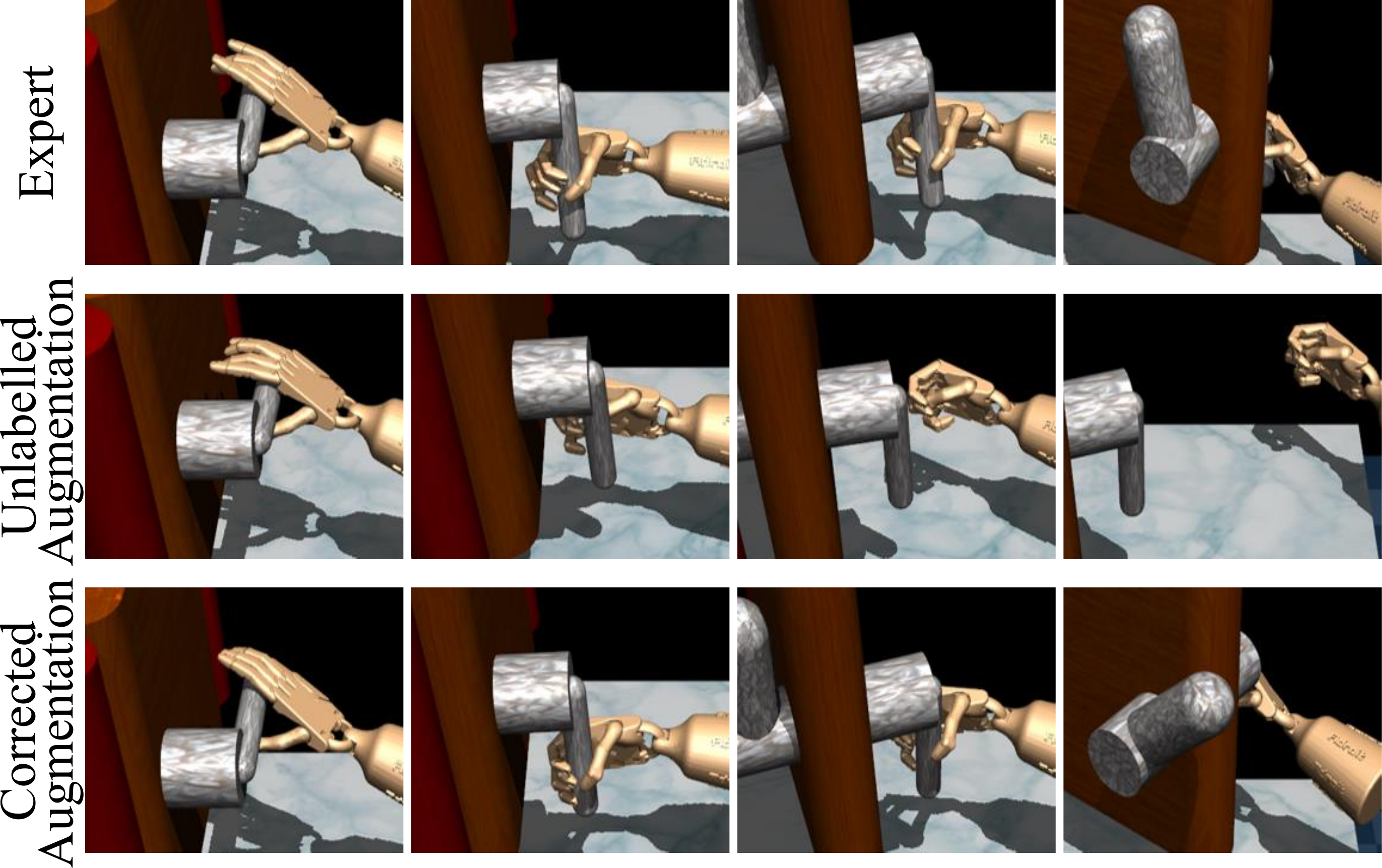}
\caption{Example of our proposed method for trajectorial data augmentation. \textbf{Top}: the original expert trajectory. \textbf{Middle}: the expert trajectory distorted by noise. The distortion makes the trajectory unsuccessful. This result is not guaranteed, therefore this augmentation is unlabelled. \textbf{Bottom}: our correction network modifies the unlabelled augmentation to produce a successful trajectory, different from the expert.}
\label{fig:aug_frames}
\end{figure}
Data augmentation is often used in computer vision tasks~\cite{krizhevsky2012imagenet,simonyan2014very,zhong2020random} by distorting the input images. This increases the size of the labelled dataset by capturing relevant perturbations (e.g. translations, rotations). However, data augmentation in control cannot use random distortions because they can greatly affect the state of the environment. Instead, it is usually done for specific behaviours~\cite{bojarski2016end,george2018imitation,buhet2019conditional}. Additionally, distortions in sequential actions lead to compounding errors that can distort the final outcome. For these reasons, simply distorting input actions on a control trajectory does not guarantee the result will remain successful. Figure~\ref{fig:aug_frames} shows an example  of this effect on a simulated task where a robot hand needs to open a door. In this example, an expert trajectory (top row) is randomly distorted with perturbed actions, leading to a distorted trajectory where the agent fails to open the door (middle row). The proposed method described in this work allows us to ``correct'' this trajectory leading to a successful one (bottom row). 

\par
As stated above, noise in the acquisition process can affect expert demonstrations, to the point they become unsuccessful in the agent's domain, making them unusable. %This is exacerbated in tasks which define success in a binary manner (i.e. success or failure).
However, these sequences still contain useful information and could potentially become successful with small corrections to their actions. This scenario is very similar to the artificial distortion of action sequences, as seen in Figure~\ref{fig:aug_frames}. Therefore, a correction mechanism can not only help augment the dataset after, but potentially during the acquisition process as well.
\par

\begin{figure}[t]
\centering
\includegraphics[width=0.49\textwidth]{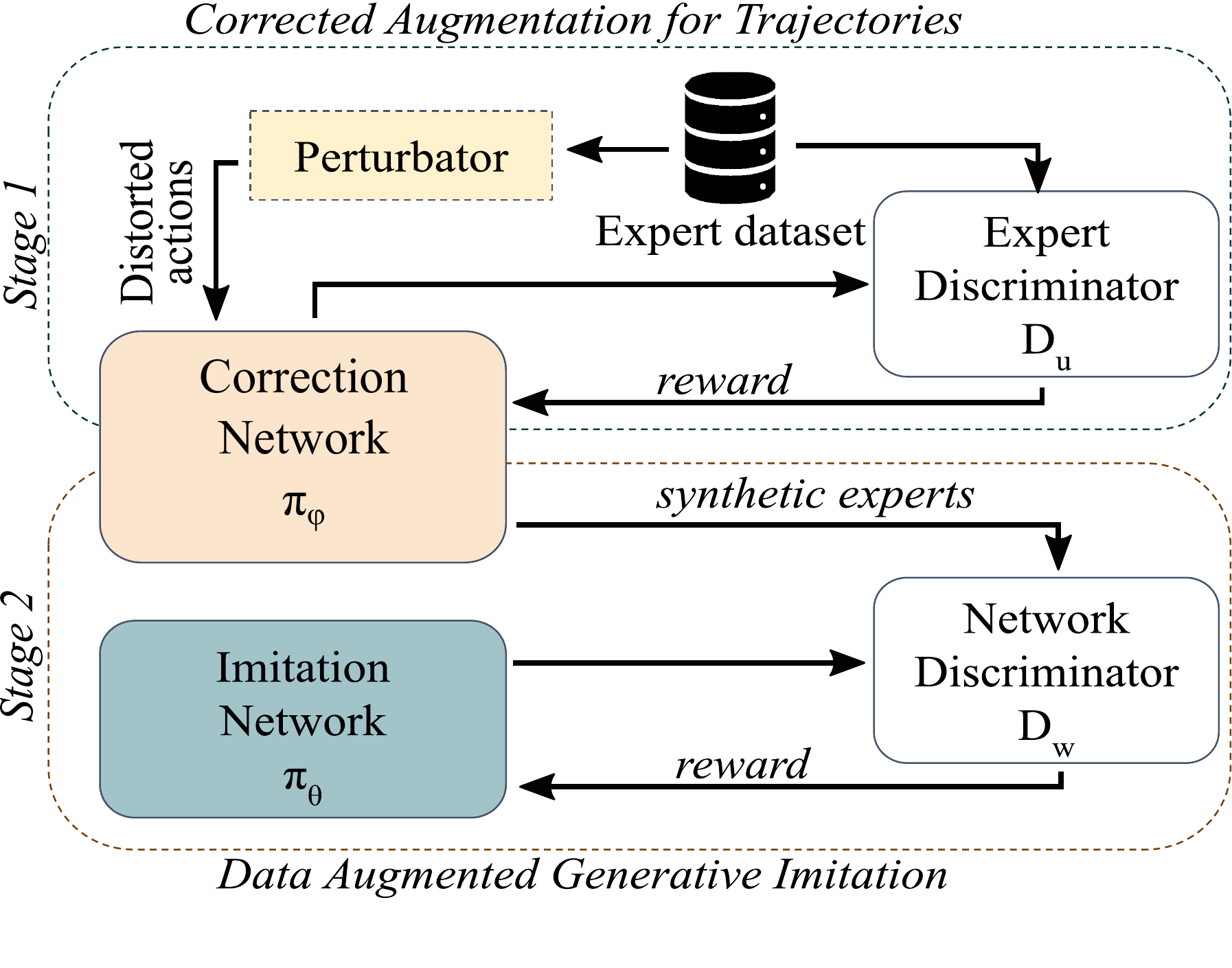}
\caption{Flow chart of our system, which performs imitation using trajectorial data augmentation. Stage 1 presents data augmentation by correcting distorted trajectories, while stage 2 presents data augmented imitation.}
\label{fig:system}
\end{figure}

The primary contribution of our work is a system that performs data augmentation on trajectories to alleviate expert demonstration acquisition and improve imitation. To achieve this, we introduce a semi-supervised adversarial framework that corrects distorted action sequences. Additionally, we present a supervised learning system which performs imitation by using a potentially infinite number of synthetic expert trajectories. Contrary to standard supervised methods which use a fixed dataset of labels, our model utilises the correction network as a synthetic expert generator. This results in a dynamic creation of experts that improve both accuracy and stability of imitation. We also introduce a novel way of measuring diversity in trajectorial datasets, in order to ensure the variance in the generated trajectories is comparable to the real expert ones. We test the correction policy's ability to correct distorted sequences, as well as its ability to help imitation by comparing the data augmented policy against the state-of-the-art GAIL and RL methods in various environments, including complex manipulation tasks with real-life experts, as presented in~\cite{Rajeswaran-RSS-18}. Our experiments show that such architecture can provide more successful corrections than random distortions, as well as improve stability and convergence over the state of the art. It also seems to retain most of the diversity of the expert trajectories without mode collapsing, thus indicating that random dynamic experts can offer a better representation of the state-action space. A general overview of our framework is depicted in Figure~\ref{fig:system}.

%-------------------------------------------------------------------------
%%%%%%%%%%%%%%%%%%%%% RELATED WORK %%%%%%%%%%%%%%%%%%%

\section{RELATED WORK}
\par{\bf Generalisation in IL} Like many other areas of Machine Learning, IL has benefited widely from deep learning~\cite{hussein2017imitation,ho2016generative,wang2017robust,li2017infogail,rahmatizadeh2017vision,schroecker2017state, songmulti}. Deep learning, though, usually requires a large amount of training data to be successful, due to its non-linearity~\cite{krizhevsky2012imagenet,simonyan2014very}. One of its most famous IL methods is Behavioural Cloning (BC)~\cite{pomerleau1989alvinn}, which is easy to train but suffers from compounding errors and cannot generalise to unseen states~\cite{ross2011reduction}. In an effort to alleviate this,~\cite{ho2016generative} introduced an approach called generative adversarial imitation learning (GAIL), which was inspired by generative adversarial networks (GANs)~\cite{goodfellow2014generative}, applied to control tasks. In their method,~\cite{ho2016generative} sought to combine imitation with RL~\cite{schulman2015trust,heess2015learning} in an effort to uncover a robust policy without the difficult task of defining an RL reward function. The generalisation problem, though, is still an issue for GANs~\cite{thanh2019improving}, and is exacerbated in GAIL due to the limited number of experts available in control problems. One reason behind GAIL's inability to generalise is due to the increased diversity of the expert trajectories. Both~\cite{wang2017robust} and~\cite{li2017infogail} tackled this problem through the use of a latent space that was able to interpolate between trajectories.

\par{\bf Few-shot IL and Semi-supervision} Since the acquisition of many expert demonstrations is difficult and not even always possible, there have been many efforts to successfully use deep learning with few~\cite{wang2018low}, one~\cite{duan2017one,finn2017one}, and even no experts~\cite{xian2017zero,peng2017zero}.~\cite{duan2017one} combined one-shot learning with meta-learning~\cite{thrun1998learning} so that only the information related to the current action and task would be used. While~\cite{duan2017one} focused on generalising a task based on its length and number of stages,~\cite{finn2017one} focused on generalising based on different settings. Similarly to our work,~\cite{wang2018low} used an additional network that produced synthetic samples using noise, but in an image classification scenario. However, their objective is different than ours, since they improved the classifier, whereas we are interested in the generated samples themselves. Another work related to ours, but in an RL scenario, is the one in~\cite{liu2019self}. In this work,~~\cite{liu2019self} designed an RL self-teaching method in an adversarial architecture, whereas our work focuses on supervised learning and specifically teaching using data augmentation. Teacher-student (T/S) networks are also used in semi-supervised setting with either noisy labels or unlabelled data~\cite{qi2019small}. In a recent work,~\cite{zhang2018deep} introduced a student cohort architecture. Similarly to our model, it aims to match the output distribution of two networks, but, unlike our work, all the students perform the same task. In terms of correcting noisy labels,~\cite{veit2017learning} used a correction network, but only to remove the noise, while we aim to make the trajectories successful after perturbation.

\par{\bf Data augmentation} Data augmentation is the practice of transforming the original labels in order to increase the size of the dataset and to describe better the label space. This practice is widely used in computer vision problems, such as image classification~\cite{krizhevsky2012imagenet,simonyan2014very,zhong2020random}. In control, it has mostly been applied to tasks that use images as input, especially in the autonomous driving setting~\cite{bojarski2016end,george2018imitation,buhet2019conditional}.~\cite{george2018imitation} applied data augmentation by distorting the images and then providing the necessary labels for these instances. Similarly,~\cite{bojarski2016end} emulated erratic behaviour and provided the necessary correction actions. The work that is mostly related to ours is~\cite{buhet2019conditional}, where they added random noise without prior knowledge of the corrective action. However, they did identify reasons of failure after the perturbation and then provided corrections. The main difference between all these works and ours is they either model the perturbation knowing its result, or involve an expert in the loop to indicate the appropriate corrections. Our work, on the other hand, has no such knowledge or resources and instead uses an adversarial method to match the expert distribution unassisted. 

In terms of generative data augmentation,~\cite{Bowles2018GANAA} applied it in image segmentation datasets. It concluded that, provided there is enough information in the original limited dataset, the generated images boost the performance of image segmentation. As for evaluating the performance between generated and real samples,~\cite{Tanaka2019DataAU} showed synthetics have similar and sometimes even better performance than the original ones, which supports our main motivation that synthetic data augmentation can improve performance.
%GAN\cite{Bowles2018GANAA,nielsen2019gan,Tanaka2019DataAU,sandfort2019data}
%%%%%%%%%%%%%%%%%%%%% FRAMEWORK %%%%%%%%%%%%%%%%%%
\section{PROPOSED METHOD}

{\bfseries Background.} We consider a Markov Decision Process, consisting of the tuple $\left(\boldsymbol{S},\boldsymbol{A},\boldsymbol{P},\boldsymbol{R}\right)$, with $\boldsymbol{S}=\left\{s_1,s_2,\dotsc\right\}$ a set of states, $\boldsymbol{A}=\left\{a_1,a_2,\dotsc\right\}$ a set of actions, $\boldsymbol{P}\left(s^\prime | s,a\right)$ the probability that action $a$ at state $s$ will lead to state $s^\prime$ and $\boldsymbol{R}\left(s,a\right)$ is a reward that evaluates action $a$ at state $s$ according to the task. We aim to learn a policy $\pi\left(a|s\right)$ describing the probability of taking an action $a$ when the agent is in state $s$, with the ideal goal to maximise the cumulative reward $\boldsymbol{R}$ across an entire trajectory (i.e. the return). Whereas the reward function needs to be defined in RL, imitation infers it from demonstrations instead. Additionally, while Inverse Reinforcement Learning aims to directly infer the reward function~\cite{ng2000algorithms}, GAIL uses a discriminator policy that provides $\boldsymbol{R}$ based on its similarity to the experts. A demonstration dataset is presented as a set of trajectories $\boldsymbol{\mathcal{T}}=\left\{\tau_1,\tau_2,\dotsc\right\}$, where each trajectory $\tau=\left\{\left(s_1,a_1\right),\left(s_2,a_2\right),\dotsc\right\}$ is a sequence of state-action pairs. 

{\bfseries GAIL.}
\begin{comment}
{\footnotesize
\begin{algorithm}[t]
% local change
\SetKwInput{KwData}{Input}
\DontPrintSemicolon
\SetAlgoLined
\KwData{expert policy $\pi_E$, initial imitation policy and network discriminator parameters $\theta_0$, $w_0$}
 \For{$i = 0, 1, 2, \dotsc$}{
     \textup{Sample trajectories $\tau_i \sim \pi_{\theta_i}$ \;
     $w_{i+1} \leftarrow \; w_i + \nabla_w L_{w_i}$ using equation~(\ref{eq:gail-d})\;
     $\theta_{i+1} \leftarrow \;  \theta_i + \nabla_\theta L_{\theta_i}$ using equation~(\ref{eq:gail-g}) and the TRPO rule\;
     }
 }
 \caption{GAIL Algorithm}
 \label{algo:gail}
\end{algorithm}
}
\end{comment}
Our architecture is based on the adversarial model of GAIL~\cite{ho2016generative}, which combines supervised classification with RL. Similarly to GANs, the policy network $\pi_\theta$ (with parameters $\theta$) generates trajectories trying to fool a discriminator $D_w$ (with parameters $w$), while the latter tries to distinguish between the real expert policy $\pi_E$ and the generated samples. Therefore, the discriminator's loss is
% \begin{equation}
% \min_{\theta} \max_{w} \mathbb{E}_{\pi_{E}} \left[\log D_{w}(s, a) \right] + 
%   \mathbb{E}_{\pi_{\theta}}\left[\log (1- D_{w}(s, a)) \right],
% \end{equation}
% where $\pi_E$ is the expert policy, even though in reality it describes a fixed expert dataset $\boldsymbol{\mathcal{T}}_E$ and $\mathbb{E}_{\pi_{E}}$ is an approximation of its state-action pair expectation.
% \par
% The discriminator's role is to act as a binary classifier between $\pi_E$ and $\pi_\theta$ and reward the samples accordingly. Ideally, it wants to give a reward of 1 to $\pi_E$ samples, thus maximising $\mathbb{E}_{\pi_{E}} \left[\log D_{w}(s, a) \right]$, while giving 0 to $\pi_\theta$ samples, thus minimising $\mathbb{E}_{\pi_{\theta}}\left[\log (1- D_{w}(s, a)) \right]$. This would maximise the difference between the two.
% \end{comment}
\begin{equation}
L_{w}= -\mathbb{E}_{\pi_E}\left[\log D_w(s,a)\right] - \mathbb{E}_{\pi_\theta}\left[\log (1-D_w(s,a))\right],
\label{eq:gail-d}
\end{equation}
while the generator's loss is
\begin{equation}
   L_\theta=\mathbb{E}_{\pi_{\theta}}\left[\log (1- D_{w}(s, a)) \right]. 
   \label{eq:gail-g}
\end{equation}
In order to get the expectation of $D_w$ with respect to $\pi_\theta$, GAIL models it as an RL cost function and approximates it with a gradient such as TRPO~\cite{schulman2015trust}.

\subsection{Framework Overview}
Our framework, presented in Figure~\ref{fig:system}, is separated into two stages. In stage 1, we introduce a way to augment trajectory datasets. That is achieved through an adversarial architecture that performs Corrected Augmentation for Trajectories (CAT), which aims to correct actions of experts that have been distorted. In stage 2, we use the CAT network of stage 1 to dynamically generate synthetic experts, in order to train a Data Augmented Generative Imitation (DAugGI) agent.
\par
CAT's objective is to produce successful demonstrations from experts. The main difference between this semi-supervised model and standard imitation is the fact that this network has access to noisy sequences of expert actions and its objective is to correct them, as shown in the example in Figure~\ref{fig:aug_frames}. Due to the stochastic nature of the policy, the resulting semi-supervised network can produce an infinite amount of corrected trajectories, which can then be used to augment small datasets of expert trajectories.
\par
The DAugGI network leverages these augmented datasets to perform imitation learning. More precisely, rather than using a limited expert dataset (such as GAIL), we use CAT to dynamically produce synthetic experts.  Ideally, by having access to a much larger (and potentially infinite) set of experts, DAugGI should be significantly faster and more stable to train.

%%%%%%%%%%%%%%%%%%%%% CAT %%%%%%%%%%%%%%%%%%%%%%%

\subsection{Corrected Augmentation for Trajectories (CAT)}
\begin{figure}[t]
\centering
\includegraphics[width=0.47\textwidth]{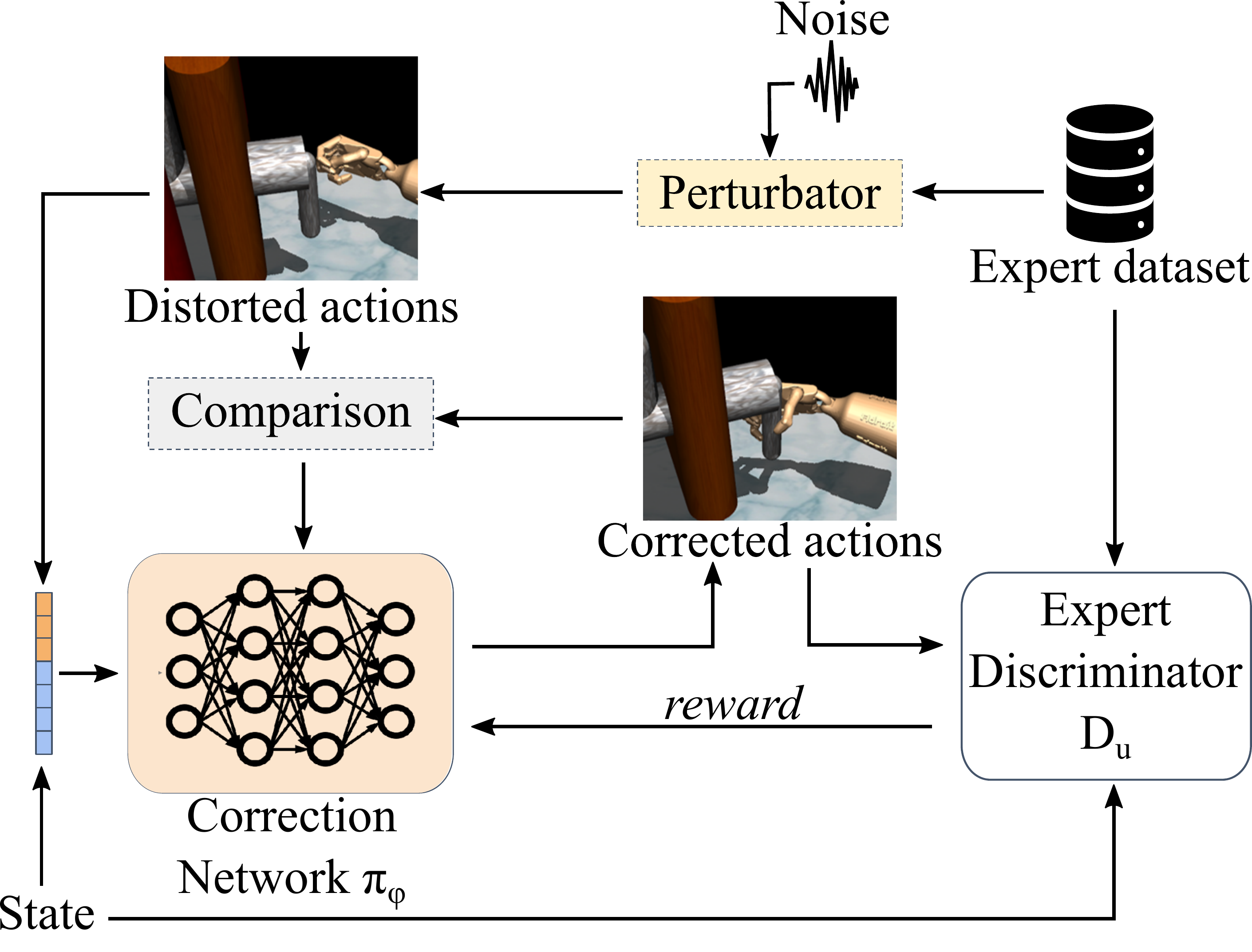}
\caption{Detailed overview of stage 1, which performs Corrected Augmentation For Trajectories. The architecture is semi-supervised since it is guided by unlabelled distorted actions.}
\label{fig:semi}
\end{figure}
Using data augmentation in supervised approaches increases the diversity of the samples and the dataset's coverage of the input space. The benefits of this regularisation technique are a better robustness and an improved resistance to overfitting~\cite{zhang2017understanding}. It is difficult to perform data augmentation in control, though, due to the fact that distorting the input (i.e. adding noise to the actions) can very easily change the label of the trajectory (i.e. it becomes unsuccessful when the original was not). Finding the right amount of noise that preserves the labels can be arduous and limiting in terms of diversity in the dataset.
%Examples of this behaviour are shown in Figure~\ref{fig:fails}. 
Therefore, data augmentation through action distortion in control can only be performed if there is a guarantee the label does not change.
\par
To tackle this, we propose a semi-supervised CAT framework which aims to correct naively distorted expert actions, thus producing new successful synthetic demonstrations. As shown in Figure~\ref{fig:semi}, CAT's input is the environment's state $s$ concatenated with distorted expert actions $a^\prime$ from distorted expert action sequences
\begin{equation}
\begin{gathered}
    q = \left\{a_{1}^\prime,a_{2}^\prime,a_{3}^\prime,\dotsc\right\}, \quad
     \textup{where} \quad a_{t}^\prime = a_{E_t} + \nu \quad \\ \textup{and} \quad \tau_E = \left\{(s_{E_1},a_{E_1}),(s_{E_2},a_{E_2}),\dotsc\right\}.
     \label{eq:tu}
\end{gathered}
\end{equation}
The distorted actions $a'$ are produced by perturbing the expert actions $a_E$ of expert trajectories $\tau_E$ with uniform noise $\nu$ (with standard deviation $\sigma$).
%These distorted action sequences are based on expert trajectories $\tau_E$, the actions $a_E$ of which are perturbed by uniform noise $\nu$ (with standard deviation $\sigma$), producing distorted actions $a'$. %These are then used as auxiliary state.

\begin{comment}
\begin{figure}[t]
\centering
\includegraphics[width=1\textwidth]{fails5.pdf}
\caption{Perturbed trajectories in different environments that became unsuccessful due to the added noise.}
\label{fig:fails}
\end{figure}
\end{comment}
{\footnotesize
\begin{algorithm}[t]
% local change
\SetKwInput{KwData}{Input}
%\SetKwInput{KwResult}{Output}
\SetKwComment{Comment}{}{}
\DontPrintSemicolon
\SetAlgoLined
\KwData{Set of expert trajectories $\mathcal{T}_E$, regularisation $\lambda$ noise $\sigma$, initial policy $\phi_0$ and discriminator $u_0$, $N$ number of perturbed action sequences.}
%\KwResult{\(particle_{global}^{best}\)}
\Comment*[l]{// produce randomly perturbed augmented action sequences $\boldsymbol{\mathcal{Q}}$}
$\boldsymbol{\mathcal{Q}} = \{\}$\;
% \For{$\textup{\textbf{each}}~\tau_{i}~\textup{\textbf{in}}~\mathcal{T}_E$}{
%     \textup{create $N$ perturbed action sequences $\boldsymbol{\mathcal{Q}}_{u_i} = \left\{q_{u_{i_1}},\dotsc,q_{u_{i_N}}\right\}$ according to~(\ref{eq:tu})}\;
%     $\boldsymbol{\mathcal{Q}}_u = \left\{\boldsymbol{\mathcal{Q}}_u, \boldsymbol{\mathcal{Q}}_{u_i}\right\}$\;
%  }
 \For{$\textup{\textbf{each}}~\tau_E~\textup{\textbf{in}}~\mathcal{T}_E$}{
    \textup{Generate $N$ perturbed action sequences $\boldsymbol{\mathcal{Q}'} = \left\{q_1,\dotsc,q_N\right\}$ from $\tau_E$ according to~(\ref{eq:tu})}.\;
    $\boldsymbol{\mathcal{Q}} = \boldsymbol{\mathcal{Q}} \bigcup\boldsymbol{\mathcal{Q}}'$.\;
 }
 \Comment*[l]{// correct augmented trajectories so they are successful}
 \For{$i = 0, 1, 2, \dotsc$}{
     \textup{Sample trajectory $\tau_i \sim \pi_{\phi_i}(q_i)$, with $q_i$ sampled uniformly from $\boldsymbol{\mathcal{Q}}$.\;
      $u_{i+1} \leftarrow \; u_i - \nabla_u L_{u_i}$ using GAIL's~(\ref{eq:gail-d}).\;
      $\phi_{i+1} \; \leftarrow \;  \phi_i - \nabla_\phi L_{\phi_i}$ using~(\ref{eq:semi-g}).\;
     }
 }
 \caption{Corrected Augmentation for Trajectories Algorithm}
 \label{algo:semi}
\end{algorithm}
}
The resulting states of the distorted trajectories suffer from compounding errors, therefore they cannot be used as a representation of a successful demonstration. The distorted actions $a_t^\prime$, on the other hand, present a valid action sequence $q$ and their noise does not depend on the noise of previous steps. Hence, they can be used to represent an approximation of an expert action sequence. The aim of the CAT policy network $\pi_\phi(a|s,a^\prime)$ is to produce successful action sequences by following these sequences of distorted actions and minimising the difference $|a-a^\prime|$ between the two. The form of these corrected trajectories is the following:
\begin{equation}
\begin{gathered}
        \tau_c = \left\{(s_{E_1}, a_{c_1}),(s_{c_2},a_{c_2}),(s_{c_3},a_{c_3}),\dotsc\right\}, \\ \textup{ where}~a_{c_t} \sim \pi_\phi(s_{c_{t}},a^\prime_t).
\end{gathered}
\end{equation}
\begin{comment}
Additionally, the stochasticity of generated actions adds randomness to the ways an unlabelled trajectory is corrected, thus increasing the diversity of the corrected trajectories.
\end{comment}
\par
% We model CAT with an adversarial architecture since a BC strategy might be more prone to overfitting.
The loss function for the correction generator is defined as follows:
\begin{equation}
L_{\phi}= \mathbb{E}_{\pi_\phi}\left[\log(1-D_u(s,a))\right]+ \lambda ||a-a^\prime||_2^2.
\label{eq:semi-g}
\end{equation}
The main differences between~(\ref{eq:semi-g}) and GAIL's~(\ref{eq:gail-g}) are the policy's input and the addition of the second term, which utilises the unlabelled actions $a^\prime$. This alters the objective of the generator to minimising the difference between its actions and distorted expert action sequences, in addition to maximising the discriminator reward. Therefore, $a^\prime$ are used as auxiliary state. The discriminator's objective is to separate the generated samples from the fixed real experts. Since the unlabelled data are not part of its objective, its loss remains the same as GAIL's, which is presented in~(\ref{eq:gail-d}).
\begin{comment}
\begin{equation}
\begin{split}
L_{u}= & -\mathbb{E}_{\pi_E}\left[\log D_u(s,a)\right] \\ & - \mathbb{E}_{\pi_\phi}\left[\log (1-D_u(s,a))\right].
\end{split}
\label{eq:semi-d}
\end{equation}
\end{comment}
A description of the CAT process is presented in Algorithm~\ref{algo:semi}.
\par
\begin{comment}
Whereas generally adding more terms to the generator's loss can worsen its stability, in our case it amplifies it since it forces it early on to follow an accurate, albeit noisy, representation of expert action sequences. This intuition is supported experimentally in Section~\ref{sec:exp}.
\end{comment}

%%%%%%%%%%%%%%%%%%%%%% DAugGI %%%%%%%%%%%%%%%%%%%%%
\subsection{Data Augmented Generative Imitation (DAugGI)}
\begin{figure}[t]
\centering
\includegraphics[width=0.47\textwidth]{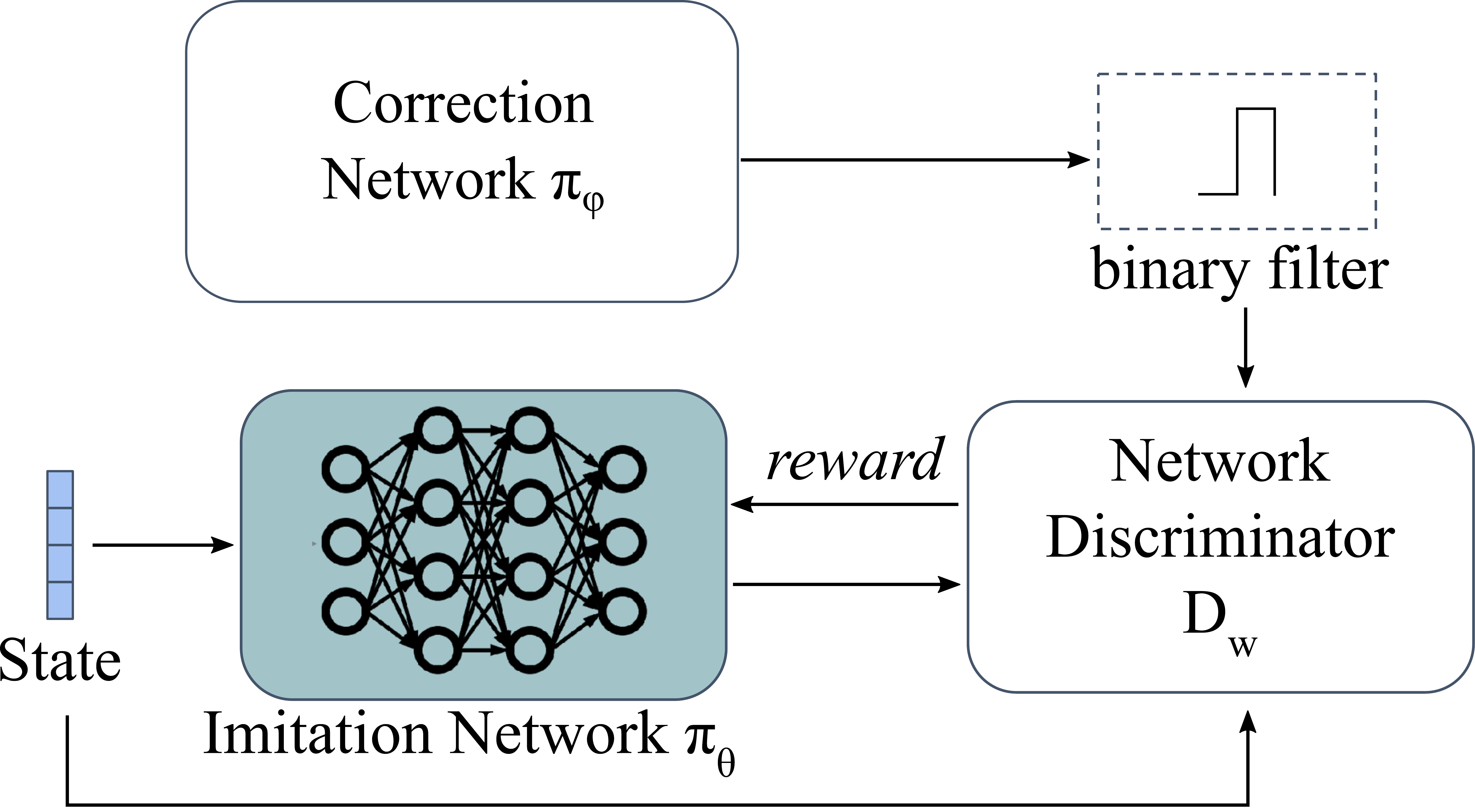}
\caption{Detailed overview of stage 2, which performs Data Augmented Generative Imitation, including the success filtering mechanism.}
\label{fig:dauggi}
\end{figure}
The ultimate goal of trajectorial data augmentation is to improve upon imitation. Data augmentation methods usually produce a fixed augmented dataset. However, CAT has already modelled the distribution of the corrected state-action pairs. Therefore, similarly to~\cite{kurenkov2019ac}, we can use the guided CAT policy to produce dynamic experts, as Figure~\ref{fig:dauggi} shows, instead of sub-sampling its distribution in a fixed dataset.
\par
Even though our method uses the output of CAT to teach an imitation network, it is not a Teacher/Student (T/S) network. Indeed, usually T/S networks have the same objective~\cite{zhang2018deep,tarvainen2017mean} (or at least same type of input~\cite{meng2018adversarial}). They can therefore share weights, at least up to a point, since they try to learn similar features. In contrast, in our setting the two networks have different objective -- one is to correct, the other is to imitate -- and different inputs. However, they do share the same output space and their resulting trajectories are comparable. That is why, in our architecture, the connection between the two networks is only in the output space, where we use the CAT network to dynamically produce synthetic experts which, in turn, teach a Data Augmented Generative Imitation (DAugGI) network to perform the task from scratch.
\par
An important difference between the corrected trajectories $\tau_c$ and the real expert ones $\tau_E$ is that the former are not guaranteed to be successful, despite the correction. Using every output of CAT to train DAugGI is beneficial only when CAT itself has a very high success rate (See Section~\ref{sec:exp}). If this is not the case, a selective mechanism is needed to filter out unsuccessful corrections so as not to confuse the generator. In most real-world applications, we usually have a-priori knowledge of the success criterion for each task and therefore we can easily use it as a filter. This information is binary and sparse, since it is provided only at the end of each trajectory and is, therefore, easy to implement in practice. Whereas its simplicity and sparsity make using it as a reward in RL very difficult, as seen in~\cite{Rajeswaran-RSS-18} as well as Section~\ref{sec:exp}, it is very beneficial in our case.
\par
The network architecture of DAugGI is similar to GAIL's, as seen in Figure~\ref{fig:dauggi}. After Stage 1, we freeze the CAT policy network and include it in the training process of the imitation network. The objective of the DAugGI policy $\pi_\theta$ is to match the expert distribution, therefore its loss function is the same as GAIL's generator, presented in~(\ref{eq:gail-g}). The discriminator $D_w$, on the other hand, tries to distinguish samples between $\pi_\phi$ and $\pi_\theta$, which are the CAT and DAugGI generators respectively, instead of $\pi_\theta$ and $\pi_E$ as in GAIL. Therefore, the loss of discriminator $D_w$ is
\begin{equation}
L_{w}= -\mathbb{E}_{\pi_\phi}\left[\log D_w(s,a)\right] -\mathbb{E}_{\pi_\theta}\left[\log (1-D_w(s,a))\right].
\label{eq:daugi-d}
\end{equation}
% Whereas using potentially infinite number of synthetic expert trajectories may make the discriminator's job more difficult, it can also safeguard against overfitting, due to the shear number of labels that are used.
\begin{comment}
An example of how the correction network and the dynamic experts can help stabilise the imitation network, even when its training is particularly difficult, can be seen in the door environment in Figure~\ref{fig:mani}.
\end{comment}
\par
Despite completing the same task, the motivation of the imitation and correction networks is distinctively different. Imitation networks, such as DAugGI and GAIL, aim to perform tasks without any supervision during the task execution. Our correction network, on the other hand, is guided during execution by an estimation of how the trajectory should be. It is then asked to micro-correct the individual actions to ensure the final outcome is a successful one (see Figure~\ref{fig:aug_frames} for an example). In terms of training, the CAT network is more stable and can learn faster, since it is given more information. This is supported by the experiments in Section~\ref{sec:exp}. In terms of diversity though, the imitation network has greater variety in trajectories, since it is free to explore the entire state-action space. The guided CAT network, on the other hand, limits its exploration around the sub-optimal trajectories and focuses around their state-action space.
\begin{comment}
This forces its trajectories to be similar to each other. Therefore, while imitation networks try to cover a big portion of the state-action space, the guided network focuses on specific parts of it.
\end{comment}
\begin{comment}
{\footnotesize
\begin{algorithm}[t]
% local change
\SetKwInput{KwData}{Input}
\SetKwRepeat{Do}{do}{while}\SetKwComment{Comment}{}{}
\DontPrintSemicolon
\SetAlgoLined
\KwData{unlabelled trajectories $\boldsymbol{\mathcal{T}}_u$, correction policy parameters $u$, initial imitation policy and network discriminator parameters $\theta_0$, $w_0$}
 \For{$i = 0, 1, 2, \dotsc$}{
  \Comment*[l]{// filter corrected trajectories so they are all successful}
    \Do{$\tau_c$ is not successful}{Sample corrected trajectories $\tau_c \sim \pi_\phi(\tau_u)$ where $\tau_{u} \sim \boldsymbol{\mathcal{T}}_u$}
     \textup{Sample trajectories $\tau_i \sim \pi_{\theta_i}$ \;
      Update the discriminator parameters $w_i$ according to GAIL with the gradient of equation~(\ref{eq:daugi-d}).\;
      Update the policy parameters $\theta_i$ using the TRPO rule with the loss function of equation~(\ref{eq:daugi-g}).}\;
 }
 \caption{Data Augmented Generative Imitation Algorithm}
 \label{algo:dauggi}
\end{algorithm}
}
\end{comment}
%%%%%%% EXPERIMENTS %%%%%%%%%%%%
\section{EXPERIMENTS}
\label{sec:exp}
This section describes the experiments conducted to evaluate the effectiveness of trajectorial augmentation in imitation. This includes the evaluation of the success and diversity of the CAT and DAugGI networks.%, as well as their diversity, through the introduction of a new diversity metric.

\subsection{Experimental Setup}
We test our method on two types of tasks. One is the classic control environments of OpenAI~\cite{1606.01540} and the other is more complex dexterous object manipulation tasks presented in~\cite{Rajeswaran-RSS-18}, which use real expert demonstrations.%All success criteria and environmental conditions can be seen in Table~\ref{tab:cond}.

\par{\bf{OpenAI Tasks}} we test our framework on two OpenAI tasks, InvertedPendulum and HalfCheetah. Whereas these two tasks were successfully trained with ample experts and environment steps in GAIL's original presentation~\cite{ho2016generative}, we increase the difficulty of the tasks by dramatically reducing the expert demonstrations to 3 and the steps of each iteration by more than 16 times. That way we test the environments in harsher conditions, where the number of resources and experts are limited. Since the CAT network manages to achieve highly successful results, no filtering mechanism is needed for the synthetic experts in DAugGI.

\par{\bf{Dexterous Object Manipulation Tasks}} these tasks involve object manipulation with a dexterous anthropomorphic hand and include a) the opening of a door, b) hammering a nail and c) pen manipulation. Due to the complexity of the tasks, all of them require binary filtering on the synthetic experts while training DAugGI. The filtering criteria used are the same as the success criteria for evaluation. All Door experiments were also pretrained with BC for 10000 iterations to speed up training.
\par 
All expert trajectories from~\cite{Rajeswaran-RSS-18} were acquired using a CyberGlove III and an HTC headset and tracker. In order to evaluate imitation with these datasets, we first train the unstable GAIL baseline, and then evaluate the effects our augmentation has on stability, success and diversity. Since we make use of a binary filter at the end of each trajectory, we also compare it to the Deep Deterministic Policy Gradient (DDPG) RL method~\cite{lillicrap2015continuous}. For a fair comparison, the DDPG agent is trained using only the binary filter as a sparse binary reward at the end of each trajectory.
\par
Regarding diversity, we test the entire original expert dataset, as well as generated datasets of 100 successful trajectories from GAIL and our two generators CAT and DAugGI. All the networks used consist of two hidden layers with 64 neurons and all the environments use $\lambda=0.1$. The iteration sizes were $2^{10}$ for InvertedPendulum and HalfCheetah, $2^{14}$ for Door and Hammer and $2^{12}$ for Pen. The simulation environment used in this work is Mujoco Pro \cite{kumar2015mujoco} and all experiments are performed using an Intel Xeon E5-2650 v2 @ 2.60GHz CPU, an NVIDIA GeForce GTX 1080 Ti GPU and 256 GB of RAM.

%%%%%%%%%%%%%%%%%%%%%%%%%%%%%%%%% CAT %%%%%%%%%%%%%%%%%%%%%%%%%%%%%%%%%%%%
\begin{figure*}[t]
\centering
\includegraphics[width=0.95\textwidth]{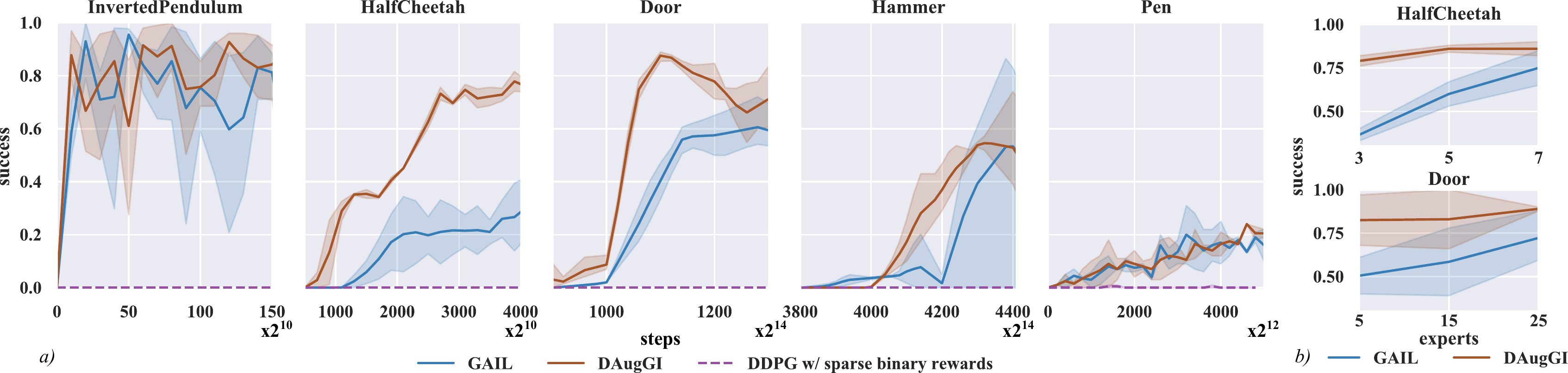}
\caption{a) Performance results of various tasks at different training steps. It includes easier OpenAI tasks (HalfCheetah and InvertedPendulum) with 3 experts, as well as more challenging dexterous manipulation tasks (Door, Hammer, Pen) with 25 experts. DAugGI is trained using the augmented CAT trajectories and generally outperforms GAIL, trained with the original limited trajectories. b) Ablation studies with different number of experts for HalfCheetah and Door tasks. DAugGI consistently outperforms GAIL, especially when the expert dataset is limited.}
\label{fig:res}
\end{figure*}

\subsection{CAT Evaluation}
\begin{table}
\caption{CAT Evaluation and Dataset Diversity}
\footnotesize
%\normalsize
\centering
\setlength\tabcolsep{2pt} % default value: 6pt
\def\arraystretch{1.2}%  1 is the default, change whatever you need
\begin{tabular}{|c||cc|c|c|cc|}
%\hline
 \multicolumn{1}{c}{} & \multicolumn{2}{c}{CAT Success \%} & \multicolumn{1}{c}{} & \multicolumn{3}{c}{Dataset Diversity score $\overline{dtw}_n$} \\
\hline
 Task & \makecell{Random\\Aug/tion} & \makecell{Corrected\\Aug/tion} & & $\dfrac{\text{CAT}}{\text{original}}$ & $\dfrac{\text{DAugGI}}{\text{original}}$ & $\dfrac{\text{GAIL}}{\text{original}}$\\ 
 \hline\hline
 HalfCheetah & 0.7 & \textbf{97.6} &  & 0.54 & 0.68 & 0.73\\ 
 \hline
 Inv. Pendulum & 4 & \textbf{100} & & 0.91 & \textbf{1.13} & 1.11 \\
 \hline
 Door &  21 & \textbf{56} & & \textbf{1.11} & \textbf{0.26} & 0.25 \\
 \hline
 Pen & 58 & 46 & & 0.93 & \textbf{1.12} & 1.06\\
 \hline
 Hammer & 62 & \textbf{63} & & \textbf{1.00} & \textbf{0.26} & 0.24\\
 \hline
\end{tabular}
\label{tab:cat}
%\end{wraptable}
\end{table}
We fist evaluate CAT by testing its ability to successfully correct the distorted action sequences. This is done by comparing the success rate of a number of corrections, compared to the success of the original random perturbations. The result of this is presented in Table~\ref{tab:cat} (left). Regarding the OpenAI tasks, CAT is almost always able to correct successfully, even under severe distortion. The dexterous manipulation tasks were more challenging, but it still managed to surpass the random perturbations in most cases. The most challenging task was the pen, potentially due to its success being more difficult to infer than the others, leading to unhelpful corrections. Still, its trajectories do not collapse since CAT's diversity is very close to the original, as seen in Table~\ref{tab:cat} (right). This indicates that even when a CAT policy is a ``bad'' teacher, it is still good enough to train a DAugGI policy successfully, as seen in Figure~\ref{fig:res}.

%%%%%%%%%%%%%%%%%%%%%%%%%%%%%%%%% DAUGGI %%%%%%%%%%%%%%%%%%%%%%%%%%%%%%%%%%%%
\subsection{DAugGI Evaluation}

We further evaluate the CAT augmentation in terms of success and stability in imitation. This is done by using CAT to train DAugGI policy networks and then comparing them to GAIL and sparse binary DDPG policies. They are evaluated by generating samples at different environmental steps, using multiple seeds, as shown in Figure~\ref{fig:res}. Due to lack of available experts, the same experts are used for all the seeds.
\par
 Figure~\ref{fig:res}~a) shows that tasks respond differently based on their difficulty. Very easy or difficult tasks, like InvertedPendulum and Pen respectively, seem to behave very similarly to GAIL. That is because the tasks are either already easily solvable (InvertedPendulum) or the ``bad'' teacher does not provide any additional information (Pen). But even then, it seems to not only retain the original information, but also increase stability. Tasks with medium difficulty, like HalfCheetah and Door, are the ones that benefited the most, with DAugGI showing clear improvement. Regarding Hammer, another medium difficulty task, DAugGI manages to greatly improve its stability. GAIL, on the other hand, proved very unstable in its ability to converge. 
\par
All the DDPG runs were unable to converge, meaning that the minimal binary success filter at the end of each trajectory is not enough information for pure RL methods. Similar findings were also reported in~\cite{Rajeswaran-RSS-18}, which used sparse rewards. Additionally, we evaluate the significance of the size of the expert dataset. Figure~\ref{fig:res} b) shows that DAugGI can improve overall performance even in extreme situations with very few experts, where GAIL typically struggles. 
%%%%%%%%%%%%%%%%%% DIVERSITY %%%%%%%%%%%%%%%

\subsection{Diversity} 
Diversity is a criterion often tested in GANs for computer vision tasks~\cite{gurumurthy2017deligan}, but it is more difficult to do so in control, due to the sequential nature of generated trajectories. In an effort to quantify the diversity capabilities of our generators, we introduce a dataset metric that utilises the Dynamic Time Warp (DTW) score between trajectories. As~\cite{vaughan2016comparing} stated, DTW offers a better metric than distance for comparison between trajectories, but can only perform pairwise comparison. In order to produce a score for an entire dataset $\mathcal{T} = \left\{\tau_1,\dotsc,\tau_N\right\}$, we calculate the mean dtw score of all the possible pairs of trajectories in a dataset and further normalise it with the diversity of the experts:
\begin{comment}
\begin{equation}
\overline{dtw}(\mathcal{T}) = \frac{\sum_{i=1}^{N-1}\sum_{j=i+1}^{N}dtw(\tau_{z_i},\tau_{z_j})}{(N-1)N/2},
\label{eq:dtw}
\end{equation}
 Since the real experts $\mathcal{T}_E$ contain the original diversity, we further normalise the diversity metric of the generators with that of the experts, so that their normalised score is:
\end{comment}
\begin{equation}
\begin{gathered}
   \overline{dtw}_n \left(\mathcal{T}_g\right)=\frac{ \overline{dtw}\left(\mathcal{T}_g\right)}{ \overline{dtw}\left(\mathcal{T}_E\right)},~\\ \textup{ where }~\overline{dtw}(\mathcal{T}) = \frac{\sum_{i=1}^{N-1}\sum_{j=i+1}^{N}dtw(\tau_{z_i},\tau_{z_j})}{(N-1)N/2}.
    \label{eq:ratio}
\end{gathered}
\end{equation}
Similarly to~\cite{brill2019exact}, $\tau_{z_i}$ is the z-normalised $\tau_i$ trajectory.
\par
The diversity ratio for each generator, using~(\ref{eq:ratio}) with FastDTW~\cite{salvador2007fastdtw}, are shown in Table~\ref{tab:cat} (right). The diversity of CAT is smaller than the other networks, which is expected, since it is guided by a plethora of similar trajectories. But we still expect it to offer a better representation of the space between the different experts. To test that, we compare the diversity results of GAIL and DAugGI, which were trained with the original experts and CAT trajectories respectively. Encouragingly, the diversity of the DAugGI network is not only very close to that of GAIL, but it even slightly surpasses it in most of the instances. That is an indication that CAT can potentially generalise slightly farther than the original expert dataset.
%%%%%%%%%%%%%%%%%%% Conclusion %%%%%%%%%%%%%%%%%%%%%

\section{CONCLUSION}
In this work, we present a data augmentation framework for control systems. Due to the nature of trajectories, it is not guaranteed that input distortion will preserve their labels. Hence, we develop a semi-supervised correction network that is guided by distorted expert actions and produces synthetic expert trajectories. Our experiments show that the correction network not only captures an at least equal, and usually better, representation of the action-space, but can also provide a faster, more stable and equally diverse training environment for imitation agents. Potential extensions of the present work are i) turning it into mutual-learning so both networks help each other, ii) incorporating the diversity metric into training, and iii) applying the framework in real-life settings, on near-expert trajectories with structured noise.

\addtolength{\textheight}{-5.5cm}   % This command serves to balance the column lengths

%%%%%%%%%%%%%%%%%%%%%%%%%%%%%%%%%%%%%%%%%%%%%%%%%%%%%%%%%%%%%%%%%%%%%%%%%%%%%%%%

% \bibliographystyle{IEEEtran}
% \newpage
% \bibliography{Ref}

%%%%%%%%%%%%%%%%%%%%%%%%%%%%%%%%%%%%%%%%%%%%%%%%%%%%%%%%%%%%%%%%%%%%%%%%%%%%%%%%

\end{document}